\documentclass[conference]{IEEEtran}
\IEEEoverridecommandlockouts
\usepackage{cite}
\usepackage{amsmath,amssymb,amsfonts}

\usepackage{textcomp}
\usepackage{xcolor}
\usepackage{graphicx}
\usepackage{url}
\usepackage{subcaption} 
\usepackage{lipsum}  
\usepackage{algorithm}
\usepackage[noend]{algpseudocode}
\usepackage{booktabs}
\usepackage{hyperref}

\def\BibTeX{{\rm B\kern-.05em{\sc i\kern-.025em b}\kern-.08em
    T\kern-.1667em\lower.7ex\hbox{E}\kern-.125emX}}
\begin{document}

\title{DRTA: Dynamic Reward Scaling for Reinforcement Learning in Time Series Anomaly Detection}

\IEEEpubid{\makebox[\columnwidth]{\hfill
2025 Conference on AI, Science, Engineering, and Technology (AIxSET)
\hfill}%
\hspace{\columnsep}\makebox[\columnwidth]{}}%
\IEEEpubidadjcol

\author{\IEEEauthorblockN{1\textsuperscript{st} Bahareh Golchin}
\IEEEauthorblockA{\textit{School of Computer Science} \\
\textit{Portland State University}\\
Portland, Oregon, United States \\
bgolchin@pdx.edu}
\and
\IEEEauthorblockN{2\textsuperscript{nd} Banafsheh Rekabdar}
\IEEEauthorblockA{\textit{School of Computer Science} \\
\textit{Portland State University}\\
Portland, Oregon, United States \\
rekabdar@pdx.edu}
\and
\IEEEauthorblockN{3\textsuperscript{rd} Kunpeng Liu}
\IEEEauthorblockA{\textit{School of Computer Science} \\
\textit{Portland State University}\\
Portland, Oregon, United States \\
kunpeng@pdx.edu}
}

\maketitle

\begin{abstract}
\label{sec:abstract}
Anomaly detection in time series data is important for applications in finance, healthcare, sensor networks, and industrial monitoring. Traditional methods usually struggle with limited labeled data, high false-positive rates, and difficulty generalizing to novel anomaly types. To overcome these challenges, we propose a reinforcement learning-based framework that integrates dynamic reward shaping, Variational Autoencoder (VAE), and active learning, called DRTA. Our method uses an adaptive reward mechanism that balances exploration and exploitation by dynamically scaling the effect of VAE-based reconstruction error and classification rewards. This approach enables the agent to detect anomalies effectively in low-label systems while maintaining high precision and recall. Our experimental results on the Yahoo A1 and Yahoo A2 benchmark datasets demonstrate that the proposed method consistently outperforms state-of-the-art unsupervised and semi-supervised approaches. These findings show that our framework is a scalable and efficient solution for real-world anomaly detection tasks.\footnote{Code is available at
  \href{https://github.com/baharehgl/Dynamic-Reward-RL-VAE}{GitHub Repository}.}
\end{abstract}


\section{Introduction}
\label{sec1:Introduction}
  
Anomaly detection in time series is crucial across various domains, including data centers, sensor networks, cyber-physical systems, healthcare, demand forecast, and finance \cite{ren2019time}, \cite{PaschalidisChen}, \cite{fontugne2013strip}, and \cite{thomas2019detecting}. Most existing algorithms in the literature require manually tuning the problem parameters and selecting features. However, this is often tailored to specific data characteristics. Additionally, identifying anomalies manually is both time-consuming and labor-intensive, and it is prone to human error. In today's world, given the vast amounts of data involved, an automated system is essential for efficiently detecting anomalies in large-scale time series datasets \cite{elaziz2023deep}.

In practice, anomaly detection faces two key challenges: first, anomalies are rare. Therefore, it is difficult to train models effectively. Second, real-world data is often time-dependent. To address the former challenge, many anomaly detection algorithms have been developed, which use unsupervised approaches that do not rely on labeled data. However, these methods are not generic. They often operate based on specific assumptions about anomaly patterns in the data, and these assumptions may not always hold in real-world scenarios, which would result in high false-positive rates. This issue stems from diverse user perspectives and varying definitions of anomalies \cite{liu2008isolation}, and \cite{zhu2017business}.

To overcome this issue, supervised methods have also been developed to detect anomalies, and they perform well when ample labeled data is available. However, they struggle in situations with limited or no labels. Even with labeled data, these algorithms assume a stable underlying distribution. When the distribution is shifted for any reasons, they require retraining to maintain accuracy \cite{wu2021rlad}.

Moreover, semi-supervised learning algorithms have also been developed to detect anomalies in time series datasets. However, these methods are limited in scenarios where diverse anomaly types are present. The reason is that semi-supervised learning algorithms rely on a small set of known anomalies, and they may fail to identify novel anomalies in unlabeled data. Consequently, they are ineffective at detecting new types of anomalies \cite{krajsic2021semi}.

To address the aforementioned challenges of weakly-supervised anomaly detection in time series data, in this paper, we propose using Deep Reinforcement Learning (DRL). A key aspect of Reinforcement Learning (RL) is the balance between exploration and exploitation \cite{sutton1998reinforcement}. Exploitation refers to selecting actions based on existing knowledge to maximize rewards. Exploration involves trying new actions to discover potentially better strategies. The agent must continuously navigate this trade-off to balance the immediate optimal decisions with the need to acquire more knowledge.

The DRL model in our study is designed to effectively use a limited set of labeled anomalous data ($D_{la}$) while extensively exploring a large pool of unlabeled data ($D_u$). This approach enables us to detect new anomaly types not present in the labeled dataset. The exploration process is improved by incorporating a Variational Autoencoder (VAE), which strengthens the DRL framework.

Additionally, given the high cost and limited availability of fully labeled data in real-world scenarios, our system uses active learning. This allows the RL agent to: 1) efficiently explore the environment and gather experience, and 2) make informed decisions based on the knowledge gained during exploration.

Finally, the core of our DRL agent is a Long Short-Term Memory (LSTM) network, which 1) models sequential time series data and 2) captures long-term dependencies between events \cite{golchinriahi}. To improve the learning efficiency of the RL agent, we use the novel idea of reward shaping, which provides additional guidance to the reinforcement learning process. By designing a well-structured reward function, we help the RL agent learn meaningful patterns more quickly. This reduces the time needed to converge to an optimal policy. Reward shaping enables the agent to receive intermediate feedback. This is particularly helpful in complex environments where rewards are sparse or delayed. This approach prevents the agent from relying only on trial-and-error exploration. It instead encourages the agent to follow a more informed learning trajectory, which ultimately improves anomaly detection performance in time series data \cite{MLVSL}.

To summarize, our key contributions include the following.
\begin{itemize}
\item \textbf{Dynamic Reward Mechanism for Adaptive Exploration and Exploitation}. This study introduces a novel dynamic reward shaping mechanism that adaptively balances exploration and exploitation, enabling the RL agent to achieve robust anomaly detection performance.

\item \textbf{Efficient Active Learning Integration with Adaptive Querying}. Our proposed method improves active learning by dynamically querying the most uncertain samples during training. This, combined with adaptive reward shaping, reduces the need for labeled data.

\item \textbf{Unified Framework for Semi-Supervised Anomaly Detection}. By integrating RL with the dynamic reward function, active learning, and VAE, this study proposes a unified framework that effectively handles low-label systems.
\end{itemize}

This paper is structured as follows: Section \ref{sec2:related} surveys relevant literature, Section \ref{sec3:background} establishes the theoretical foundation for time series anomaly detection, Section \ref{sec4:ProposedMethod} describes our proposed methodology, Section \ref{sec5:Experiment} analyzes the implementation process, and Section \ref{sec6:Conclusion} presents our conclusions.

%The structure of the rest of this paper is as follows. In Section \ref{sec2:related}, we review the studies in the literature which is related to our work. Moreover, the background in time series anomaly detection is explored in Section \ref{sec3:background}. Next, our proposed framework is detailed in Section \ref{sec4:ProposedMethod}. Section \ref{sec5:Experiment} discusses the implementation of our proposed method. Finally, we conclude our study in Section \ref{sec6:Conclusion}.

\section{Related Work}
\label{sec2:related}
In this section, we review machine learning-based approaches in the literature for time series anomaly detection.

% \subsection{Statistical-Based Anomaly Detection}
% Statistical-based methods build models from given datasets and use mathematical tests to determine whether unseen data conforms to the proposed model. These methods assume that normal data follows a specific probability distribution. Parametric models, such as Gaussian models, regression models, and logistic regression, fall into this category \cite{shekhar2001detecting}. Kernel function-based methods also learn normal behavior directly from training data \cite{kumar2016approach} and \cite{yamanishi2004online}. However, an important limitation of statistical approaches is the basic assumption that normal behavior follows an existing distribution, which often does not hold in complex, real-world scenarios.

\subsection{Machine Learning-Based Anomaly Detection}
Machine learning approaches address the limitations of statistical methods by using labeled training data to distinguish between normal and anomalous instances through classification or clustering techniques.

\textbf{Supervised and Unsupervised Learning}. Commonly used algorithms include Bayesian networks \cite{das2007detecting}, support vector machines (SVM) \cite{ma2003time}, rule-based systems \cite{tandon2007weighting}, and neural networks \cite{mukkamala2002intrusion}. Clustering techniques, such as k-means, have also been used for anomaly detection \cite{ramaswamy2000efficient}.

\textbf{Reinforcement Learning in Anomaly Detection}. More recently, RL has been used in anomaly detection due to its ability to learn from interactions with the environment. Bourdonnaye et al. proposed using convolutional autoencoders within an RL framework to detect anomalies \cite{bourdonnaye2017learning}. Huang et al. introduced a value-based DRL approach using the Deep Q-Network (DQN) algorithm \cite{huang2018towards}. Furthermore, reward shaping has emerged as an important technique in RL, particularly for enhancing anomaly detection systems. For instance, Devidze et al. introduce Exploration-Guided Reward Shaping, a self-supervised framework that accelerates reinforcement learning in sparse-reward environments \cite{RPA}. They combine learned intrinsic rewards with exploration bonuses to enhance agent performance. Eschmann et al. has a chapter specifically on how to design the reward function for the RL agent \cite{Eschmann2021}.

\subsection{Anomaly Detection in Time Series Data}
For complex time series data, recently, anomaly detection algorithms have been developed. For example, 1) Skyline \cite{esty2014skyline} is a real-time anomaly detection system, and 2) Twitter’s anomaly detection package is designed to identify anomalies in the presence of seasonality and trends \cite{twitter2015anomaly}. Some anomaly detection methods, such as ContextOSE \cite{mikhail2015contextual}, emphasize capturing local patterns rather than global trends. Hierarchical temporal memory (HTM), implemented in projects like Numenta, stores and recalls temporal and spatial patterns for anomaly detection \cite{ahmad2017unsupervised}. With the advancement of deep learning, Recurrent Neural Networks (RNNs) and Long Short-Term Memory (LSTM) networks have been widely used for predicting future values and identifying anomalies based on prediction errors. Variants of autoencoders have also been explored for anomaly detection in time series data recently \cite{malhotra2016lstm}. Perhaps, the most relevant study to ours is RLVAL \cite{BB24}. %They propose the combination of RL, VAE, and active learning in the context of anomaly detection. 
In this study, they use static rewards that cause the VAE component to dominate their reward function.

To further advance this field, a novel approach integrating the DQN algorithm with VAE, active learning, and reward shaping has been proposed in this paper. This hybrid method aims to create a more robust framework for identifying anomalies in time series data, which improves detection accuracy and adaptability.

\section{Background}
\label{sec3:background}
Before presenting our proposed method, we first provide an overview of fundamental concepts, which include 1) DQNs, 2) VAE, 3) Active Learning, and 4) reward shaping. This review ensures a clearer understanding of our approach.

\subsection{Deep Q-Networks and Q-Learning}
Q-learning learns the action-value function $Q(s, a)$ estimating expected rewards. The Q-function updates using:
\begin{equation}
Q_{k+1}(s, a) \leftarrow (1 - \alpha)Q_k(s, a) + \alpha [R_{s,a,s'} + \gamma \max_{a'} Q_k(s', a')]
\end{equation}

Traditional Q-learning with neural networks can be unstable \cite{L17}. DQNs address this using experience replay and target networks \cite{L92}. Experience replay stores transitions $<s, a, r, s'>$ to reduce sample correlation and improve efficiency. Target networks provide fixed references for stable updates and better convergence.

% \subsection{Deep Q-Networks and Q-Learning}
% One of the value-based reinforcement learning algorithms is $Q$-learning, where the agent learns the action-value function 
% $Q(s, a)$, which estimates the expected reward for taking a specific action in a given state. The target value for updates is defined as
% \begin{equation}
%    \text{target} = R_{s,a,s'} + \gamma \max_{a'} Q_k(s', a')
% \end{equation}
% Next, the following formula shows how the $Q$-function is updated.
% \begin{equation}
%    Q_{k+1}(s, a) \leftarrow (1 - \alpha)Q_k(s, a) + \alpha \cdot \text{target}
% \end{equation}

% However, it is important to note that traditional $Q$-learning can be unstable or even diverge, especially when the action-value function is approximated using nonlinear models such as neural networks \cite{L17}. To address this challenge, DeepMind introduced DQNs, which integrate reinforcement learning with deep neural networks to tackle more complex problems effectively. DQN improves action-value function approximation by using two key techniques: (1) experience replay and (2) a target network \cite{L92}. Experience replay keeps a buffer of past state transitions, each recorded as a tuple $<s, a, r, s\;'>$. 
% DQN enables the agent to learn from a diverse set of experiences, which minimizes correlations between consecutive samples, and it improves learning efficiency. The target network improves stability by providing a fixed reference for target values over a set period. This ensures smoother updates and helps the convergence of the main network.

\subsection{Variational Autoencoder}
VAEs map the original feature space to latent Gaussian distributions through an encoder--decoder architecture based on neural networks. The learning objective is to maximize the intractable marginal likelihood $p(x;\theta)$, where $x$ is a feature vector and $\theta$ represents the decoder parameters. This likelihood is approximated using the Evidence Lower Bound (ELBO)~\cite{elaziz2023deep}:

\begin{equation}
\mathcal{L}(\theta, \phi; x) = 
\mathbb{E}_{q(z \mid x; \phi)} \big[ \log p(x \mid z; \theta) \big] 
- \mathrm{KL}\!\left[q(z \mid x; \phi) \,\|\, p(z)\right],
\end{equation}

where $q(z \mid x; \phi)$ is the encoder distribution approximating the posterior, and $\mathrm{KL}[\cdot \,\|\, \cdot]$ denotes the Kullback--Leibler divergence between the approximate posterior and the prior $p(z)$.

\subsection{Active Learning in Machine Learning Systems}
Active learning enhances efficiency in machine learning systems by selectively querying unlabeled data points for expert labeling. Given a labeled dataset 
$\mathcal{L} = (X, Y)$ and an unlabeled pool 
$\mathcal{U} = \{x_1, x_2, \ldots, x_n\}$, 
a query function $Q$ identifies the most informative samples from $\mathcal{U}$ to refine a classifier $C$ while minimizing the number of labeled instances required.
One widely used strategy is Margin Sampling, which selects samples where the classifier is least confident. Specifically, it chooses the instance with the smallest margin between the top two predicted class probabilities:
\begin{equation}
x_m = \arg \min_{x \in \mathcal{U}} 
\Big( P_C(\hat{y}^1 \mid x) - P_C(\hat{y}^2 \mid x) \Big),
\end{equation}
where $P_C(\hat{y}^1 \mid x)$ and $P_C(\hat{y}^2 \mid x)$ denote the first and second most probable class predictions, respectively.

\subsection{Reward Shaping}
Reward shaping modifies the reward function in RL by incorporating domain knowledge to accelerate learning when rewards are sparse or delayed. Potential-based reward shaping (PBRS) preserves the optimal policy while adding intermediate rewards to encourage desirable behaviors \cite{hong2025adaptive}. We extend this concept by incorporating VAE reconstruction error into the reward function, detailed in Section \ref{DRL-Implementation}.

\section{Proposed Method}
\label{sec4:ProposedMethod}
\begin{figure*}[ht]
    \centering
    \includegraphics[width=0.7\textwidth]{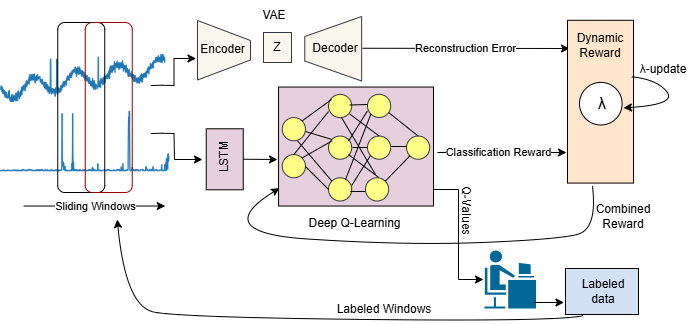}
    \caption{Workflow of our proposed method (DRTA). Input data flows through sliding windows into two parallel components: a VAE generating reconstruction error and an LSTM-based Deep Q-Learning network for classification. The Dynamic Reward component uses an adaptive coefficient $\lambda(t)$ to balance exploration (reconstruction error $R_2$) and exploitation (classification rewards $R_1$), automatically shifting focus during training. Active learning selects uncertain samples for labeling, creating an efficient feedback loop with minimal labeled data requirements.}

    \label{figure 1}
\end{figure*}

In this section, we describe our proposed method in detail. Our proposed method combines VAE, Deep RL, active learning, and reward shaping. Figure \ref{figure 1} depicts our proposed method.

\subsection{Implementing Anomaly Detection with VAE}
Because we are working with time series data, each input $x$ can be a sliding window of length $n\_steps$. By training the VAE on normal segments, the network learns a latent distribution that reflects normal behavior. During inference, we measure reconstruction error to detect anomalies. That is, higher than usual reconstruction errors can signal data points not explained well by the learned normal latent structure. In our implementation, this reconstruction error also factors into the RL agent's reward shaping, where the agent's action influences how it penalizes or rewards windows with high VAE reconstruction errors. Therefore, the VAE's role is as follows. 1) It acts as a learned feature extractor in the latent space, and 2) it provides a reconstruction-based anomaly score, which can guide the reinforcement learning policy to label anomalies more accurately.

In our proposed method, the VAE is constructed via two main components: the encoder and the decoder. We define a custom Sampling Layer that uses reparameterization, adding a noise term 
$\boldsymbol{\epsilon} \sim \mathcal{N}(\mathbf{0}, \mathbf{I})$ 
scaled by $\exp\bigl(0.5 \cdot \log \boldsymbol{\sigma}^2\bigr)$. We first build a feedforward network which outputs $\boldsymbol{\mu}(\mathbf{x})$ and $\log \boldsymbol{\sigma}^2(\mathbf{x})$. Then, the Sampling Layer combines these outputs.
After $z$ is obtained, the decoder transforms it back into the original dimensionality using a few Dense layers. The loss function is computed by summing the reconstruction error (via mean squared error or cross-entropy) and the KL term. 

\subsection{Implementing Deep RL with Reward Shaping for Anomaly Detection} \label{DRL-Implementation}

Our approach frames anomaly detection as a sequential decision-making task, where the Deep RL agent observes a window of time series data and it takes an action to classify the data as either normal or anomalous. The agent receives a reward signal that reflects: 1) detection accuracy,  and 2) additional criteria (i.e., reconstruction error from VAE). To be more specific, at each time step $t$, the environment presents the agent with a state $s_t$, which consists of a short sliding window of the time series (for instance, $n$ consecutive values). The agent's action $a_t$ is to label the last point in this window. If the agent's prediction matches the true label, it gains a positive reward; otherwise, it receives a penalty, which is a negative number. Furthermore, to encourage learning more robust latent representations, a VAE-based reconstruction error is added as a bonus or penalty in the reward function, shaping the agent's behavior to pay extra attention to regions of the time series that deviate from normal patterns. Below, we explain how we defined the state, action, policy, and reward in our proposed method.

Each \emph{state} $s_t$ corresponds to a short sliding window of length $n$ from the time series. Specifically, if the time series is $\{x_1, x_2, \dots, x_T\}$, then at time $t$, the agent observes $\{x_{t-n+1}, \dots, x_t\}$. This window provides a localized snapshot of recent data, which captures short-term trends or sudden deviations that might indicate anomalies. We focus on a limited history rather than the entire time series. By adopting this method, the state remains manageable in size and ensures that the agent can efficiently process each step for real-time or near real-time anomaly detection.

At each time step $t$, the agent takes an action $a_t \in \{0,1\}$, where $0$ indicates a prediction of normal for the last data point in the current window, and $1$ indicates a prediction of anomalous. Once the action is chosen, the environment advances by one step, which shifts the sliding window forward to form the next state $s_{t+1}$.

Next, the policy $\pi(a \mid s)$ is derived from a Q-network that estimates the action-value function $Q^\pi(s,a)$. At each time step $t$, given state $s_t$, the agent selects $a_t \;=\; \arg \max_{a \in \{0,1\}}\, Q^\pi(s_t,\,a).$ The Q-network itself is updated via the Bellman equation, which states that for an optimal policy,
\begin{equation}
Q^*(s,a) \;=\; \mathbb{E}\Bigl[r(s,a) + \gamma \,\max_{a'}\,Q^*(s',a')\Bigr],
\end{equation}
where 1) $r(s,a)$ is the reward for taking action $a$ in state $s$, 2) $\gamma \in (0,1)$ is the discount factor, and 3) $s'$ is the next state. In practice, we approximate $Q^\pi(s,a)$ with a deep neural network and iteratively minimize the temporal-difference error $\bigl(y - Q^\pi(s_t,\,a_t)\bigr)^2$.
where $y = r_t + \gamma \,\max_{a'} Q^\pi(s_{t+1},a')$ is the Bellman target. As training progresses, the agent refines its Q-function. As a result, the policy that selects normal ($a=0$) or anomalous ($a=1$) labels for each time step is also refined.

Next, we present our reward system design for semi-supervised anomaly detection in time series data, which combines immediate classification rewards (i.e., extrinsic reward, $R_1$) with reconstruction-based reward shaping through VAE integration  (i.e., intrinsic reward, $R_2$). This demonstrates superior performance in low-label systems. We define the $R_1$ structure as follows.
\begin{equation}
R_1(s_t, a_t) =
\begin{cases} 
TP_{val} & \text{if } a_t = 1 \text{ and } y_t = 1, \\ 
TN_{val} & \text{if } a_t = 0 \text{ and } y_t = 0, \\ 
FP_{val} & \text{if } a_t = 1 \text{ and } y_t = 0, \\ 
FN_{val} & \text{if } a_t = 0 \text{ and } y_t = 1.
\end{cases}
\end{equation}
where, 1) $TP_{val} = 5, TN_{val} = 1$, and 2) $FP_{val} = -1, FN_{val} = -5$. This implementation creates a reward vector where index 0 represents the reward for non-anomaly classification and index 1 for anomaly classification. The asymmetric reward structure (5:1 ratio between TP and TN) reflects the higher importance of detecting true anomalies.

%***
Next, reconstruction-based reward shaping in our proposed method uses the VAE to guide the learning process by incorporating reconstruction error as an additional reward component. The VAE is trained on normal time series data to learn a compact latent representation. This enables the VAE to reconstruct normal patterns effectively. The reconstruction error is calculated as the MSE between the original input $x_t$ and its reconstruction $\hat{x}_t$, which serves as a measure of how well the current state aligns with normal behavior. The reconstruction error is computed as:
\begin{equation}
   R_2(s_t, a_t)=\text{MSE}(x_t, \hat{x}_t) = \frac{1}{n} \sum_{i=1}^{n} (x_{t,i} - \hat{x}_{t,i})^2, 
\end{equation}
where $n$ is the dimensionality of the input window.

Finally, the mathematical formulation for the total reward is:
\begin{equation}
    R_{total}(s_t,a_t)= R_1(s_t,a_t)+\lambda(t)R_2 (s_t,a_t)
\end{equation}
where $\lambda(t)$ is a dynamic scaling coefficient that adjusts the effect of the reconstruction penalty over time. In our proposed method, the dynamic coefficient $\lambda(t)$ plays a crucial role in balancing the influence of the reconstruction error from VAE in the total reward calculation. It ensures that the agent learns to prioritize anomaly detection while still leveraging reconstruction-based guidance during training.

Reconstruction error provides an unsupervised signal that complements the supervised classification reward $R_1(s_t,a_t)$. Without scaling by $\lambda(t)$,  the magnitude of the reconstruction error might dominate or be negligible compared to $R_1(s_t,a_t)$, which could lead to suboptimal learning. By dynamically adjusting $\lambda(t)$, the framework ensures that: 1) The agent explores normal patterns early in training, and 2) the focus gradually shifts toward accurate anomaly classification as training progresses.

The coefficient (i.e., $\lambda(t)$) is updated after each episode based on the total episode reward. The update rule follows a proportional control mechanism:

\begin{equation}
\lambda_{t+1} = \text{clip}\left(\lambda_t + \alpha \left(R_{\text{target}} - R_{\text{episode}}\right), \lambda_{\text{min}}, \lambda_{\text{max}}\right),
\end{equation}

where: 1) $R_{\text{target}}$: Target reward for an episode. 2) $R_{\text{episode}}$: Total reward achieved in the current episode. 3) $\alpha$: Learning rate for adjusting $\lambda(t)$. 4) $\lambda_{\text{min}}$ and $\lambda_{\text{max}}$: Minimum and maximum allowable values for $\lambda(t)$. 5) The clip function restricts a value to stay within a specified range, replacing values below $\lambda_{\min}$ with $\lambda_{\min}$ and values above $\lambda_{\max}$ with $\lambda_{\max}$.

This formula ensures that:
\begin{itemize}
    \item If $R_{\text{episode}} < R_{\text{target}}$, then $\lambda(t)$ increases to emphasize reconstruction error.
    \item If $R_{\text{episode}} > R_{\text{target}}$, then $\lambda(t)$ decreases to reduce reliance on reconstruction error.
\end{itemize}

\begin{figure}[htbp]
    \centering
    \begin{subfigure}[b]{0.40\textwidth}
        \centering
        \includegraphics[width=\textwidth]{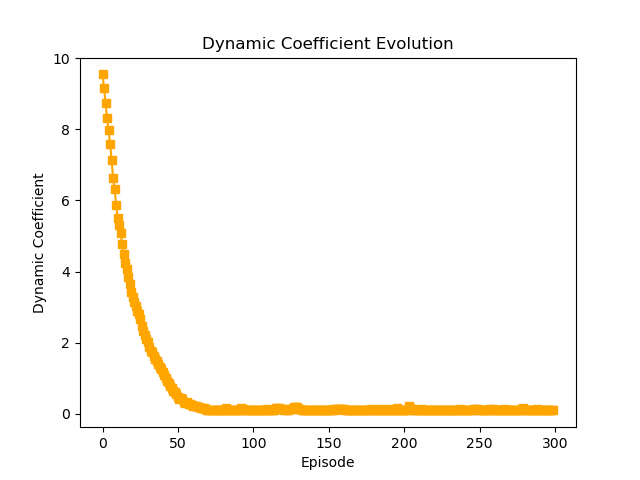}
        %\fbox{\rule[-.5cm]{0cm}{4cm} \rule[-.5cm]{4cm}{0cm}} % Replace this line with the line above, where image1 is your image.
        \caption{Dynamic coefficient evolution over episodes}
        \label{fig:sub1-coef-reward curve}
    \end{subfigure}
    \hfill
    \begin{subfigure}[b]{0.40\textwidth}
        \centering
        \includegraphics[width=\textwidth]{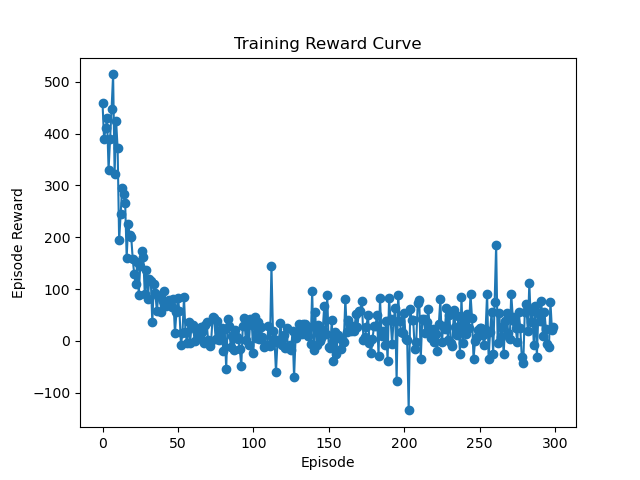}
        %\fbox{\rule[-.5cm]{0cm}{4cm} \rule[-.5cm]{4cm}{0cm}} % Replace this line with the line above, where image2 is your image.
        \caption{Training reward curve}
        \label{fig:sub2-coef-reward curve}
    \end{subfigure}
    \caption{Relationship between the dynamic coefficient and reward evolution during training.}
    \label{fig:coef-reward curve}
\end{figure}

Figure \ref{fig:coef-reward curve} depicts the relationship between the dynamic coefficient evolution and the training reward during RL training. Figure \ref{fig:sub1-coef-reward curve} shows how the scaling factor $\lambda(t)$ decreases over episodes. It starts at a high value to prioritize exploration (via reconstruction error) and gradually stabilizes as the agent shifts focus to exploitation (classification accuracy). This behavior directly affects figure \ref{fig:sub2-coef-reward curve}, where initial episodes show higher rewards due to the significant contribution of reconstruction error ($R_2$) scaled by $\lambda(t)$. As $\lambda(t)$ decreases, the reward curve stabilizes and reflects classification performance ($R_1$), with fluctuations arising from variations in state-action transitions. These figures demonstrate how the dynamic reward mechanism effectively balances exploration and exploitation throughout training.

\subsection{Implementing Active Learning for Anomaly Detection}
In our proposed method, the active learning module is designed to iteratively identify and label the most uncertain samples from the time series dataset to improve the agent's anomaly detection capabilities. The active learning class uses a margin-based sampling strategy. It calculates the absolute difference between the Q-values of the two possible actions (normal or anomaly) for each state (i.e., $\text{Margin}(s) = |Q(s, a_1) - Q(s, a_2)|$). First, samples with the smallest margin (i.e., the most uncertain predictions) are ranked. Then, the top-N uncertain samples are selected for manual labeling by a user (i.e., $\text{Selected Samples} = \arg \min_{s \in \mathcal{S}} \text{Margin}(s)$). This process ensures that the agent focuses on learning from the most ambiguous cases. This accelerates the agent's capability in detecting anomalies.

Once these samples are labeled, they are added back to the dataset. Then, label propagation is applied using a semi-supervised learning technique (i.e., LabelSpreading) to transmit labels to nearby unlabeled samples based on feature similarity. The probability of a label $y_i$ for an unlabeled sample $x_i$ is computed as:

\begin{equation}
   P(y_i | x_i) = \frac{\sum_{j \in \mathcal{L}} w_{ij} P(y_j | x_j)}{\sum_{j} w_{ij}} 
\end{equation}
 where 1) $\mathcal{L}$ is the set of labeled samples, and 2) $w_{ij}$ is the similarity weight between samples $x_i$ and $x_j$. This combination of active learning and label propagation: 1) reduces reliance on large amounts of labeled data, and 2) enables the agent to learn effectively. The active learning process is tightly integrated into the RL loop, which allows the agent to refine its policy progressively by incorporating high-value labeled samples into its training set. Our proposed method algorithm is detailed in Algorithm \ref{alg:rl-anomaly}.

\begin{figure}[htbp]
    \centering
    \begin{subfigure}[b]{0.40\textwidth}
        \centering
        \includegraphics[width=\textwidth]{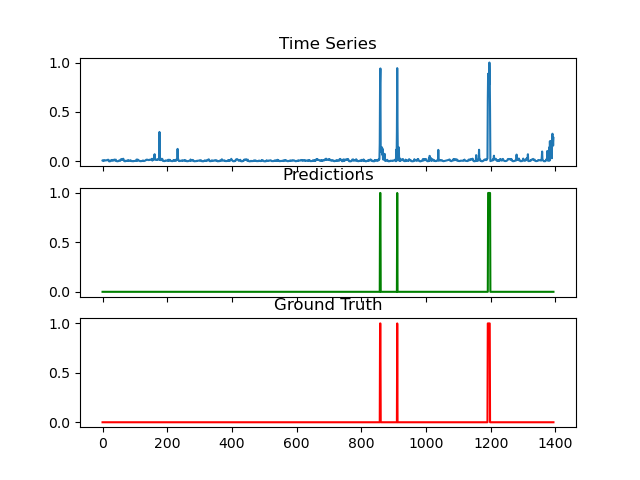}
        %\fbox{\rule[-.5cm]{0cm}{4cm} \rule[-.5cm]{4cm}{0cm}} % Replace this line with the line above, where image1 is your image.
        \caption{An episode for Yahoo A1 dataset with 1\% queried samples}
        \label{fig:sub1}
    \end{subfigure}
    \hfill
    \begin{subfigure}[b]{0.40\textwidth}
        \centering
        \includegraphics[width=\textwidth]{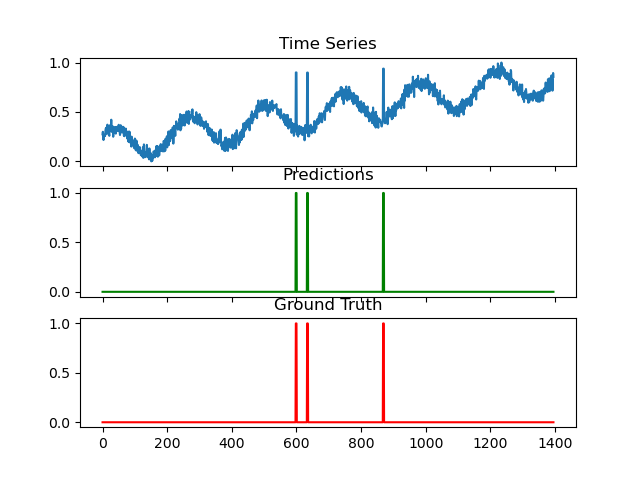}
        %\fbox{\rule[-.5cm]{0cm}{4cm} \rule[-.5cm]{4cm}{0cm}} % Replace this line with the line above, where image2 is your image.
        \caption{An episode for Yahoo A2 dataset with 10\% queried samples}
        \label{fig:sub2}
    \end{subfigure}
    \caption{Example visualizations of anomaly detection results from the proposed method.}
    \label{fig:subfigureExample}
\end{figure}

Figure \ref{fig:subfigureExample} illustrates the performance of our proposed method on two different datasets and query rates. Figure \ref{fig:sub1} represents an episode from the Yahoo A1 dataset with 1\% of samples queried via active learning, which shows how well our model aligns its predictions with the ground truth anomalies. Figure \ref{fig:sub2} depicts an episode from the Yahoo A2 dataset with 10\% of samples queried, which highlights the model's ability to detect anomalies effectively even with a higher query rate and a different dataset.

\begin{algorithm}[htbp]
  \small
  \resizebox{\columnwidth}{!}{%
    \begin{minipage}{\columnwidth}
      \caption{Dynamic Reward–Scaled RL with VAE and Active Learning}
      \label{alg:rl-anomaly}
      \begin{algorithmic}[1]
        \Require Normal-data path $D$, episodes $N$, batch size $B$, initial coefficient $\lambda_0$
        \Ensure Trained Q-network; precision/recall/F1

        \Function{BuildVAE}{$D, n\_steps$}
          \State Load sliding windows from $D$ and standardize
          \State Train VAE on normal windows
          \State \Return $\mathtt{vae}$, encoder
        \EndFunction

        \Function{WarmUp}{$env, init\_mem$}
          \State Fit IsolationForest on initial states; collect top-$M$ outliers
          \State Play random/heuristic actions to fill $\mathit{ReplayMem}$
          \State \Return $\mathit{ReplayMem}$
        \EndFunction

        \Function{TrainRL}{$env, \mathtt{vae}, \lambda, N, B$}
          \State Init Q-network \& target network
          \For{$e \gets 1$ \textbf{to} $N$}
            \State Reset env; total\_reward $\gets 0$
            \While{not done}
              \State Select action via $\epsilon$-greedy on Q-network
              \State Observe $(s,r,s',\mathit{done})$
              \State $r \gets r_{\mathrm{class}} + \lambda \times r_{\mathrm{VAE}}$
              \State Append $(s,r,s',\mathit{done})$ to $\mathit{ReplayMem}$
              \State Sample mini-batch of size $B$; update Q-network
              \If{step mod $K = 0$}
                \State Sync target network
              \EndIf
            \EndWhile
            \State Perform active learning: query top-$k$ uncertain states
            \State $\lambda \gets \text{update\_dynamic\_coef}(\lambda,\text{total\_reward})$
          \EndFor
        \EndFunction

        \Function{Validate}{$env, \text{Q-network}, episodes$}
          \State Run episodes with $\epsilon = 0$; accumulate precision/recall/F1
        \EndFunction

        \Function{Main}{$D, n\_steps, init\_mem, N, B$}
          \State $\mathtt{vae}$, encoder $\gets$ \Call{BuildVAE}{$D, n\_steps$}
          \State Initialize env
          \State $\mathit{ReplayMem} \gets$ \Call{WarmUp}{$env, init\_mem$}
          \State \Call{TrainRL}{$env, \mathtt{vae}, \lambda_0, N, B$}
          \State \Call{Validate}{$env, \text{Q-network}, \lceil N/10 \rceil$}
          \State Save reward \& coefficient curves
        \EndFunction
      \end{algorithmic}
    \end{minipage}%
  }
\end{algorithm}

\section{Experiments}
\label{sec5:Experiment}
This section outlines our experiments. We first begin with dataset specifications, and then present comparative analyses against benchmark methods. We evaluate anomaly detection efficacy through three common metrics: 1) Precision measures prediction accuracy via correctly identified anomalies, 2) Recall assesses system sensitivity through true anomaly detection rates, and 3) F1-Score harmonizes both measures to mitigate evaluation bias.
%The hyperparameters used in our method are mentioned in Appendix \ref{sec:appendix1}.
 
\subsection{Datasets}
We evaluated our proposed method using two widely used datasets for time series anomaly detection: the Yahoo A1Benchmark and A2Benchmark. Table \ref{tab1} summarizes the datasets' characteristics, with detailed descriptions provided in the following section.

\paragraph{Yahoo Benchmark.}
This Yahoo Webscope benchmark combines real-world service metrics with synthetic examples for temporal anomaly detection. We use two subsets: A1Benchmark (67 labeled time series, 1400–1600 points each) from Yahoo login patterns and synthetic scenarios, and A2Benchmark (100 synthetic time series with single outlier anomalies). Both provide precise anomaly annotations for strict method evaluation.
% Developed by Yahoo's Webscope team for anomaly detection in temporal data, this benchmark combines real-world service metrics with artificially generated examples. Our analysis focuses on two subsets: the A1Benchmark, which contains 67 labeled time series (1400–1600 points each) derived from Yahoo member login patterns and synthetic scenarios. The A2Benchmark, which includes 100 synthetic time series (1400–1600 points each) with anomalies represented as single outliers. Both datasets provide precisely annotated anomalies at each temporal observation point which enables strict evaluation of anomaly detection methods.

%\paragraph{KPI Dataset.}
%Sourced from industry-academic collaborations in AIOps research, this comprehensive repository aggregates operational metrics from Tencent, eBay, and Sogou infrastructure systems. Its 3 million timestamped entries document multidimensional server performance indicators across varying load conditions. Each record contains annotated normal/abnormal states verified through post-incident root cause analyses, offering granular insights into production environment failures.

\begin{table}[ht]
\caption{Overview of datsets}
    \centering
    \begin{tabular}{|c||c|c|}
        \hline
         dataset & total points & anomalies\\
         \hline
        Yahoo A1 & 94866 & 1669\\
        \hline
        Yahoo A2 & 142100 & 400 \\
        \hline
    \end{tabular}
    \label{tab1}
\end{table}

\subsection{Results and Discussions}
In this section, we compare our proposed method to the best studies in the literature. These studies are as follows. 1) Luminol, which is a lightweight Python library that uses various statistical anomaly detection algorithms to directly return anomaly scores without requiring training \cite{linkedin_luminol}. 2) SPOT, which detects anomalies in streaming univariate time series by dynamically selecting thresholds based on extreme value theory. 3) DSPOT extends SPOT by adapting thresholds dynamically to changes in data distribution \cite{zhu2002learning}. 4) SR-CNN, which generates synthetic training data with injected fake anomalies and applies Spectral Residual (SR) methods to train its neural network\cite{ren2019time}. 5) DAGMM combines a deep autoencoder for low-dimensional representation with a Gaussian Mixture Model (GMM) to jointly optimize reconstruction error and probabilistic modeling \cite{zong2018deep}. 6) Autoencoder uses replicator neural networks with three hidden layers to measure the outlyingness of data records based on reconstruction error \cite{hawkins2002outlier}. 7) Deep SAD learns a latent distribution of normal data with low entropy while ensuring that anomalies have higher entropy distributions \cite{pang2019deep}. 8) RLAD, which combines RL and active learning to detect anomalies \cite{wu2021rlad}. 9) RLVAL, which improves RLAD by using VAE \cite{BB24}.

The results in Table \ref{tab:yahoo_A1} and Table \ref{tab:yahoo_A2} demonstrate the effectiveness of our proposed method in anomaly detection across the Yahoo A1 and Yahoo A2 datasets, in particular when compared to both unsupervised and semi-supervised baselines. Among the unsupervised methods, SPOT and DSPOT achieve relatively higher F1-scores (e.g., 0.446 and 0.442 on Yahoo A1, respectively). However, their performance is significantly lower than semi-supervised methods that leverage labeled data. Semi-supervised methods such as Deep SAD and RLAD show improved performance as the number of queried samples increases (from 1\% to 5\% to 10\% of total data samples), with RLVAL achieving an F1-score of 0.921 on Yahoo A1 when querying 10\% of samples. However, our proposed method outperforms all baselines at lower query rates, which achieves an F1-score of 0.90 on Yahoo A1 and 0.80 on Yahoo A2 with only 1\% queried samples. This demonstrates our proposed method's robustness in low-label systems.

At 5\% queried samples, our method achieves an F1-score of 0.888 on both Yahoo A1 and Yahoo A2 datasets, maintaining competitive performance compared to RLVAL (F1-score: 0.872 on Yahoo A1). This trend continues at 10\%, where our method achieves consistent results (F1-score: 0.666), while RLVAL  outperforms on Yahoo A1 (F1-score: 0.921). Notably, RLVAL does not report results for Yahoo A2, making it difficult to assess its generalizability across datasets. In contrast, our method demonstrates strong performance across both datasets, which highlights its adaptability.

It is crucial to note that the adaptive reward mechanism in our proposed method plays an important role in reducing the need for labeled data while maintaining high anomaly detection performance. At 1\% queried samples, the dynamic reward shaping mechanism effectively balances exploration (via reconstruction error, $R_2$) and exploitation (via classification rewards, $R_1$), which enables the model to achieve high recall (1.00 on Yahoo A2 and 0.90 on Yahoo A1) while maintaining precision. This balance allows our method to outperform RLVAL at this query rate (F1-score: 0.834 on Yahoo A1), which demonstrates its ability to generalize from minimal supervision.

As the number of queried samples increases to 5\% and 10\%, the adaptive reward mechanism shifts focus toward classification accuracy by dynamically adjusting the contribution of reconstruction error through $\lambda(t)$. This ensures that precision improves without sacrificing recall, enabling consistent performance across datasets with minimal dependency on labeled data. The integration of active learning further improves this efficiency by focusing labeling efforts on the most uncertain samples, which ensures that limited labeling resources are utilized effectively.

In summary, our proposed method achieves superior performance across various query rates by using adaptive rewards and active learning to reduce labeling requirements while maintaining robust anomaly detection capabilities across diverse datasets.

% -----------------------
% Table for Yahoo A1
% -----------------------
\begin{table}[htbp]
  \small
  \centering
  \caption{Performance comparison on Yahoo A1}
  \label{tab:yahoo_A1}
    \begin{tabular}{l|ccc}
      \toprule
      \textbf{Models}  & \textbf{F1-score} & \textbf{Precision} & \textbf{Recall} \\
      \midrule
      \multicolumn{4}{l}{\textbf{Unsupervised}} \\
      \midrule
      \hline
      luminol       & 0.177 & 0.261 & 0.258 \\
      SR-CNN        & 0.254 & 0.408 & 0.382 \\
      SPOT          & 0.446 & 0.513 & 0.394 \\
      DSPOT         & 0.442 & 0.512 & 0.398 \\
      DAGMM         & 0.295 & 0.317 & 0.309 \\
      Autoencoder   & 0.438 & 0.471 & 0.431 \\
      \midrule
      \multicolumn{4}{l}{\textbf{Semi-supervised}} \\
      \midrule
      \hline
      Deep SAD (1\%)  & 0.594 & 0.518 & 0.689 \\
      Deep SAD (5\%)  & 0.671 & 0.603 & 0.681 \\
      Deep SAD (10\%) & 0.727 & 0.672 & 0.758 \\
      \hline
      RLAD   (1\%)    & 0.708 & 0.652 & 0.652 \\
      RLAD   (5\%)    & 0.752 & 0.710 & 0.800 \\
      RLAD   (10\%)   & 0.797 & 0.733 & 0.922 \\
      \hline
      RLVAL  (1\%)    & 0.834 & 0.819 & 0.850 \\
      RLVAL  (5\%)    & 0.872 & 0.846 & 0.900 \\
      RLVAL  (10\%)   & \textbf{0.921} & 0.894 & 0.950 \\
      \midrule
      \multicolumn{4}{l}{\textbf{Proposed Method}} \\
      \midrule
      \hline
      DRTA (1\%)  & \textbf{0.900} & 0.900 & 0.900 \\
      DRTA (5\%)  & \textbf{0.888} & 1.000 & 0.800 \\
      DRTA (10\%) & 0.666 & 1.000 & 0.500 \\
      \bottomrule
    \end{tabular}%
\end{table}

% -----------------------
% Table for Yahoo A2
% -----------------------
\begin{table}[htbp]
  \small
  \centering
  \caption{Performance comparison on Yahoo A2}
  \label{tab:yahoo_A2}
    \begin{tabular}{l|ccc}
      \toprule
      \textbf{Models}  & \textbf{F1-score} & \textbf{Precision} & \textbf{Recall} \\
      \midrule
      \multicolumn{4}{l}{\textbf{Unsupervised}} \\
      \midrule
      \hline
      luminol       & 0.277 & 0.314 & 0.266 \\
      SR-CNN        & 0.374 & 0.408 & 0.346 \\
      SPOT          & 0.691 & 0.681 & 0.646 \\
      DSPOT         & 0.595 & 0.625 & 0.467 \\
      DAGMM         & 0.455 & 0.458 & 0.459 \\
      Autoencoder   & 0.465 & 0.484 & 0.435 \\
      \midrule
      \multicolumn{4}{l}{\textbf{Semi-supervised}} \\
      \midrule
      \hline
      Deep SAD (1\%)  & 0.754 & 0.712 & 0.799 \\
      Deep SAD (5\%)  & 0.825 & 0.801 & 0.854 \\
      Deep SAD (10\%) & 0.851 & 0.832 & 0.879 \\
      \hline
      RLAD   (1\%)    & \textbf{1.000} & 1.000 & 1.000 \\
      RLAD   (5\%)    & \textbf{1.000} & 1.000 & 1.000 \\
      RLAD   (10\%)   & 1.000 & 1.000 & 1.000 \\
      % RLVAL rows removed
      \midrule
      \multicolumn{4}{l}{\textbf{Proposed Method}} \\
      \midrule
      \hline
      DRTA (1\%)  & 0.800 & 0.666 & 1.000 \\
      DRTA (5\%)  & 0.888 & 0.800 & 1.000 \\
      DRTA (10\%) & \textbf{1.000} & 1.000 & 1.000 \\
      \bottomrule
    \end{tabular}%
\end{table}

\section{Conclusion}
\label{sec6:Conclusion}
In this study, we proposed a novel RL-based framework for time series anomaly detection which integrates dynamic reward shaping, VAE, and active learning. Our method effectively balances exploration and exploitation by dynamically scaling reconstruction error contributions through the adaptive coefficient. This enables the agent to learn robust anomaly detection policies even in low-label systems. The integration of active learning further improves efficiency by focusing labeling efforts on the most uncertain samples. This significantly reduces the dependency on large labeled datasets. Experimental results on the Yahoo A1 and A2 datasets demonstrate that our method consistently outperforms state-of-the-art approaches. For future work, we suggest exploring the integration of large language models (LLMs) into our framework to enhance interpretability and decision-making in anomaly detection.

%\bibliography{main}
%\bibliographystyle{plain}
% \newpage

\end{document}